\documentclass[letterpaper, 10 pt, conference]{ieeeconf}  
\IEEEoverridecommandlockouts                              
\usepackage{graphicx} 
\usepackage[export]{adjustbox}
\usepackage{amsmath} 
\usepackage{subcaption}
\usepackage{arydshln}
\usepackage{comment}
\usepackage[skip=2pt]{caption}
\usepackage{url}
\usepackage{multirow}
\usepackage{booktabs}
\usepackage[noadjust]{cite}

\usepackage[dvipsnames]{xcolor}

\newcommand\TODO[1]{\textbf{\textcolor{red}{#1}}}
\newcommand\UPDATED[1]{{\textcolor{blue}{#1}}}

\title{\LARGE \bf
Explicitly incorporating spatial information to recurrent networks for agriculture
}

\author{Claus Smitt, Michael Halstead, Alireza Ahmadi, and Chris McCool
\thanks{The authors are with the University of Bonn, Gernamy; Agricultural Robotics department
{\tt\small \{csmitt, michael.halstead, alireza.ahmadi, cmccool\}@uni-bonn.de}}
}

\begin{document}

\maketitle
\thispagestyle{empty}
\pagestyle{empty}


\begin{abstract}

In agriculture, the majority of vision systems perform still image classification.
Yet, recent work has highlighted the potential of spatial and temporal cues as a rich source of information to improve the classification performance. 
In this paper, we propose novel approaches to explicitly capture both spatial and temporal information to improve the classification of deep convolutional neural networks. 
We leverage available RGB-D images and robot odometry to perform inter-frame feature map spatial registration.
This information is then fused within recurrent deep learnt models, to improve their accuracy and robustness.
We demonstrate that this can considerably improve the classification performance with our best performing spatial-temporal model (ST-Atte) achieving absolute performance improvements for intersection-over-union (IoU[\%]) of $4.7$ for crop-weed segmentation and $2.6$ for fruit (sweet pepper) segmentation.
Furthermore, we show that these approaches are robust to variable framerates~and odometry errors, which are frequently observed in real-world applications.

\end{abstract}

\section{Introduction}
\label{sec:intro}

In recent years, the adoption of robotic systems in agriculture has seen a significant surge.
This uptake has been driven by recent state-of-the-art research to automate tasks such as weeding~\cite{Bawden17_1}, yield estimation~\cite{smitt2021} and harvesting~\cite{lehnert2017autonomous,arad2020development}.
Key to these advances has been the novel vision systems to enable scene understanding and crop monitoring.

Crop monitoring is the ability to locate objects of interest in the scene and provide important information to the robot; such as species-level classification or fine-grained locations.
This is a fundamental component of any platform operating in the agricultural domain, from arable farmlands to glasshouses.
The ability to accurately classify and locate weeds or crops in a scene is an invaluable tool for agricultural and horticultural robots.
Computed indices such as yield, ripeness, and size of the fruit generates important information to the farmer creating more dynamic decision making.
Without this information, other tasks such as monitoring and harvesting cannot be successfully performed in real-world situations.

In agriculture, the majority of vision systems perform still image classification.
From the early work of Nuske et al.~\cite{Nuske_2011_6891} and even the more recent deep convolutional neural network (DCNN) approach of Sa et al.~\cite{sa2016deepfruits}.
It has only been recent work that highlighted the potential of spatial and temporal cues as a rich source of information to improve the outputs and classification results of DCNNs~\cite{halstead2021crop, alireza2021Virtual, lottes2020jfr} applied to agriculture. 
Despite these advances, it remains an open question about how best to combine the spatial and temporal information to perform better classification in agricultural environments.


\begin{figure}[t!]
    \centering
    \includegraphics[width=\columnwidth]{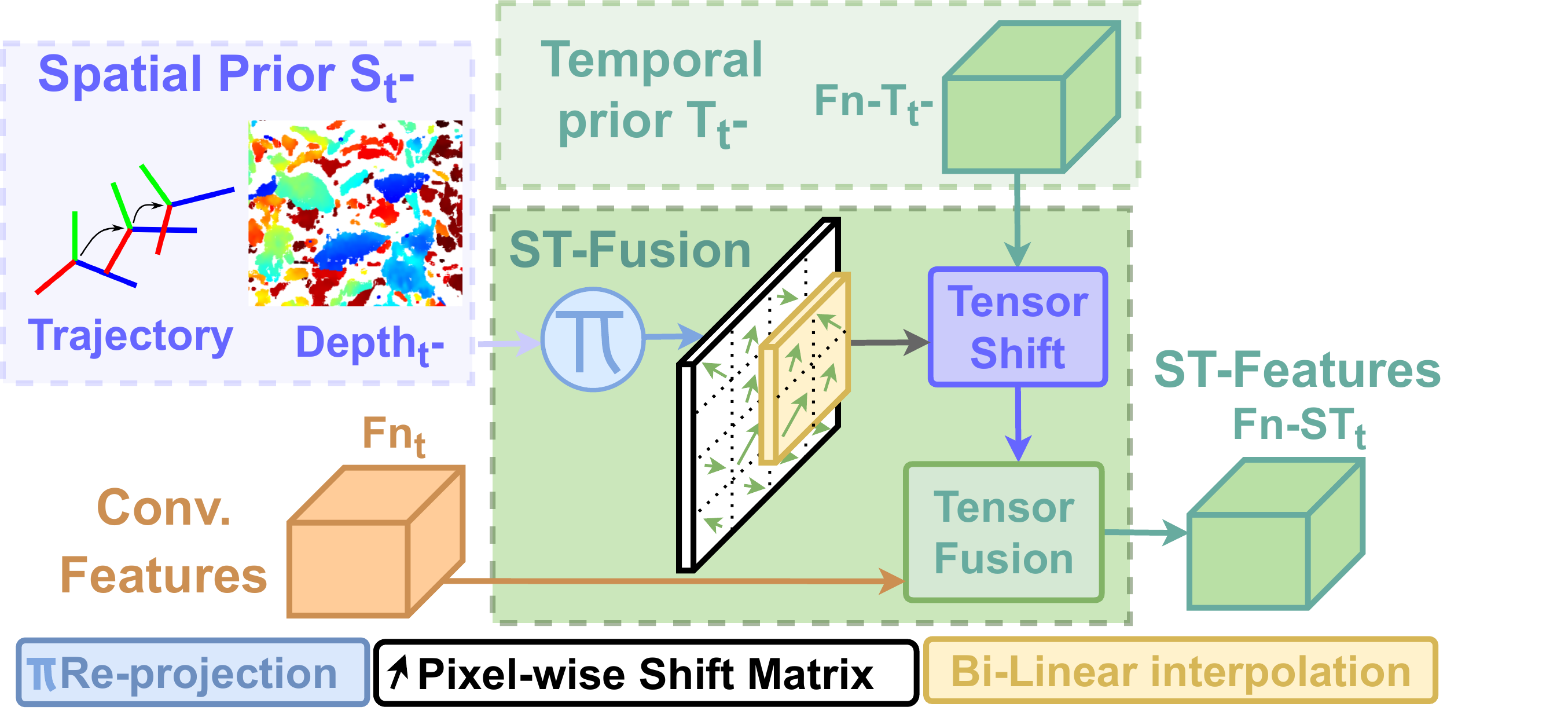}
    \caption{Overview of our proposed spatial-temporal (ST) fusion layer which combines spatial ($S_{t^{-}}$) and temporal ($T_{t^{-}}$) priors with current features ($F_n{t}$) using the Tensor registration and Fusion modules.}
    \label{fig:fusion}
    \vspace{-16pt}
\end{figure}


In this paper, we propose approaches to capture both spatial and temporal information to improve the performance of DCNNs, with our spatial-temporal fusion layer (Fig. \ref{fig:fusion}) at their core.
We leverage the available RGB-D images and robot odometry to perform inter-frame feature map spatial registration.
This information is then fused within recurrent, or temporal deep learnt models, to improve the accuracy and robustness of semantic segmentation.
In doing so we make the following contributions:
\begin{itemize}
    \item two end-to-end semantic segmentation pipelines that explicitly incorporate spatial-temporal information;
    \item A suite of validation experiments on two challenging agricultural scenarios gathered by different robots; and
    \item A robustness analysis of these methods under typical variable framerate and noisy robot-pose situations.
\end{itemize}

\section{Related Work}
\label{sec:related}

Robotic platforms are beginning to operate in diverse agricultural environments to automate key tasks. 
These include automating weeding, yield estimation, and harvesting.
A key component for all of these tasks is their vision systems that enable them to accurately and efficiently make sense of their environment.
Below we provide a brief review of such systems and the approaches they enable.

\begin{figure*}[!t]
    \vspace{12pt}
    \centering
    \includegraphics[width=0.9\textwidth]{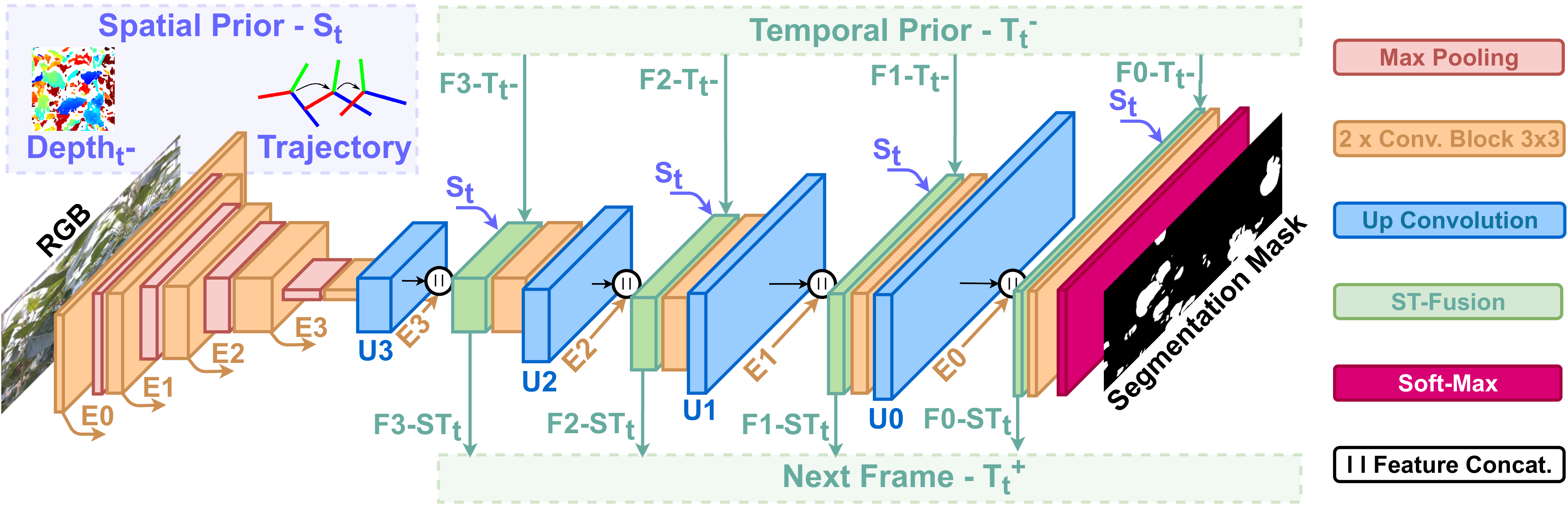}
    \caption{General diagram of our spatial-temporal segmentation architecture with the proposed ST-Fusion layers explicitly incorporating spatial and temporal priors.}
    \label{fig:pipeline}
    \vspace{-16pt}
\end{figure*}

\subsection{Traditional vision based crop monitoring}

Localization of objects in a field is key for monitoring approaches to succeed.
Nuske et al.~\cite{Nuske_2011_6891} concentrated on yield prediction for grapes and proposed the use of radial basis symmetry along with other traditional machine learning (ML) and computer vision (CV) approaches.
To calculate the yield of tomatoes~\cite{Yamamoto:2014aa} used RGB cameras along with artificial light at night to ensure consistency, using different color spaces classified by a pixel-wise decision tree.
Bawden et al.~\cite{Bawden17_1} introduced a robotic weeding platform for plant-specific weeding.
For detection, they applied color thresholding using multivariate Gaussian models based on the chrominance of different color spaces on RGB-NIR images.
Traditional CV and ML techniques were used by~\cite{utstumo2018robotic}, to control a drop-on-demand weeder that could increase the dosage for weeds only.
They employed support vector machines using both color (RGB and HSV) and shape information, along with connected components for leaf separation.

\subsection{DNN based vision and monitoring systems}
More recently, deep learning (DL) approaches have dominated research due to their ability to improve state-of-the-art results.
For fruit detection tasks, Faster-RCNN~\cite{ren2015faster} is a commonly used network, both in agriculture and horticulture settings, where~\cite{sa2016deepfruits} was one of the earliest to adapt it for this purpose.
Faster-RCNN was also compared to Yolo~\cite{Redmon_yolo_16} for mango~\cite{koirala2019deep} and apple~\cite{tian2019apple} detection, where Yolo performed with higher accuracy in real-time.
Wan and Goudos~\cite{wan2020faster} augmented Faster-RCNN to achieve real-time inference, while~\cite{halstead2018fruit} introduced a parallel layer that improved ripeness level classification without losing generalised performance~\cite{halstead2020}.
For strawberry detection~\cite{kirk2020b} used a multi RBG-D camera array and fused the RGB and LAB images in a RetinaNet architecture for promising results. 
In 2020, \cite{khan2020ced} proposed a modified U-Net~\cite{ronneberger2015u} architecture with separate pathways for weed and crop segmentation and, as well as~\cite{hashemi2022deep}, compared results with variants of the popular SegNet~\cite{badrinarayanan2017segnet} and DeepLabV3~\cite{chen2017rethinking} models in different agricultural scenarios. Both found that the performance comparisons are strongly dependent on the data domain.
More recently, extensions to DCNNs that incorporate temporal and spatial-temporal cues have been considered.

\subsection{DCNNs exploiting spatial and temporal information}

Limited work has been conducted on exploiting spatial and temporal information for improved DCNN performance in agriculture.
Lottes et~al.~\cite{lottes2020jfr} proposed to learn segmentation and detection tasks from image temporal sequences, leveraging the regular planting intervals of the crop as an implicit spatial prior, however, such planting interval does not exist in horticultural applications.
In contrast, our approach leverages both explicit spatial and temporal information, removing the assumptions about crop/fruit layouts.
Ahmadi et~al.~\cite{alireza2021Virtual} augmented a U-Net structure to create a recurrent neural network.
Their model achieved strong results for semantic segmentation in agriculture and in this work we enhance their approach by proposing recurrent networks that explicitly incorporate robot motion to improve performance.
In other application domains, the fusion of RGB and depth (spatial) information has been considered by using a GRU structure for object recognition~\cite{Loghmani2019} and~\cite{valada2019self} used a pixel-wise self-attention mechanism for multi-modal fusion.
We explore both of these approaches to fuse spatial-temporal information and employ ConvGRUs~\cite{ballas2015delving} to maintain spatial consistency through time.
Similarly, for segmentation,~\cite{xiang2017darnn} used depth information for inter-frame spatial feature map pixels association for a modified RNN decoder layer, effectively incorporating temporal spatially registered information.
Our methods extends this concept to multi-resolution vectorized registration in a DCNN which is end-to-end trainable.
To the best of our knowledge, our method is one of the first works that explicitly incorporates spatial and temporal cues in an end-to-end deep learning framework for agriculture.

\section{Proposed Approach}
\label{sec:methods}

In our approach, we exploit the spatial and temporal information that is widely available from robotic platforms to considerably improve the performance of still-image models.
We present two novel approaches to explicitly incorporate the available 3D spatial information into deep recurrent neural networks, exploiting the spatial-temporal information present in video sequences.
At their core, they take spatially registered feature maps from the previous decoder frame ($t^{-}$) (See Sec.~\ref{sec:tensorReproj}) as an input to the current frame ($t$).
The difference between these models is how they further process this information to treat it as spatial-temporal attention (see Sec.~\ref{sec:models}).
We refer to this as spatial-temporal fusion.

An overview of our proposed spatial-temporal architecture and its application to a recurrent neural network (RNN) structure is depicted in Fig.~\ref{fig:pipeline}.
For the RNN structure, we propose to take the prior information ($t^{-}$) from the decoder layer and transform it using the spatial-temporal module given in Fig.~\ref{fig:fusion}.
Finally, a critical element in the spatial-temporal architecture is registered scene geometry (3D information) which we obtain from an RGB-D camera and robot odometry.

These novel models are deployed in two challenging agricultural environments: arable farming (crop vs weed) and horticulture (fruit segmentation).
An advantage of evaluating in these environments is that the scene (crop or fruit) remains relatively static as the robot moves through the scene.
This avoids the need to disambiguate movement within the scene and enables us to provide insights into the pros and cons of our proposed approaches.
Furthermore, the data that we use consists of challenging outdoor scenes with high levels of occlusion and is gathered by a robot with RGB-D sensors.

\subsection{Feature map spatial registration}
\label{sec:tensorReproj}
We propose a novel spatial-temporal module that exploits scene geometry and estimated robot motion to extend standard deep recurrent neural networks (e.g. RNNs or GRUs).
In particular, we explicitly incorporate prior spatial information by fusing feature maps from the prior frame $\mathbf{F}_{t^{-}}$ and the current frame $\mathbf{F}_{t}$.
This is achieved by exploiting scene geometric cues to spatially register these feature maps to a common feature plane.
In this work, we register the prior map $\mathbf{F}_{t^{-}}$ to the current one $\mathbf{F}_{t}$ to produce $\mathbf{F}^{*}_{t^{-}}$.
To simplify the notation $i$ represents $t^{-}$ and $j$ represents $t$.




In general, to register a feature map $\mathbf{F}_i$ to frame $j$, we compute the camera homogeneous transform $Tc_{ij}$ by using the robot transform between them $Tr_{ij}$. 
We also account for the camera extrinsics to the robot's coordinate frame $T_{e}$ as follows, 
\begin{equation}
    Tc_{ij} = T_{e}^{-1} Tr_{ij} T_{e}.
\end{equation}
Given a depth image of size $W\times H$ and a feature map $\mathbf{F}_i$ of size $W\times H \times C$, where $C$ is its number of channels and $\mathbf{q}_{ik}=(u_k,v_k)$ is the image coordinate of the feature element value $f_{k}$, $\mathbf{F}_{i}$ can be registered into frame $j$ as follows,

\begin{subequations}
\vspace{-5mm}
\label{eq:tensorReproj}
\begin{align}
    \mathbf{q}_{jk} = \pi(Tc_{ij}(\pi^{-1}(\mathbf{q}_{ik},d_{q_{ik}}))) \label{eq:coordReproj} \\[6pt]
    \mathbf{F}^{*}_{j}(\mathbf{q}_{jk},c_k) \ = f_{jk}^{*} = f_{ik} =\  \mathbf{F}_{i}(\mathbf{q}_{ik},c_k) \label{eq:valueReproj}
\end{align}
\end{subequations}
where $\pi(.)$ is the pin-hole camera model projection function, $d_{q_.}$ is each map coordinate's depth and $T(.)$ applies a homogeneous transform to a 3D point.
This way, each feature map element $f_{ik}$ gets spatially shifted in the image plane but stays in the same feature channel $c_k$, maintaining its original meaning.
To avoid potential numerical issues, such as too small or too large gradients during training, the registered feature map $\mathbf{F}^{*}_{j}$ is initialized with a small value (i.e. 0.1).
Those points which will be re-projected outside the current feature map $\mathbf{F}_{i}$ are discarded.
FCNNs have the property of translation equivariance, which means that these transforms can be applied directly to their feature maps whilst preserving spatial consistency in the image plane.

%

\subsection{Variable size feature map registering}

 Fusing spatial-temporal information from different depths of a DCNN  implies coping with feature maps of varying sizes.
Consider  that we have a depth image of size $W\times H$, we can use Eq.~\ref{eq:coordReproj} to compute a re-projection shift matrix $\Delta_{ij}$ of size $W\times H\times 2$ for all pixels in the image, with its $k$-th element being the shift vector,
%
\begin{equation}
\vspace{-1mm}
\label{eq:shiftMatrix1}
    \Delta_{ij}(\mathbf{q}_{jk}) =  \boldsymbol{\delta}_{\mathbf{q}k} = \mathbf{q}_{jk} - \mathbf{q}_{ik}.
\end{equation}

If we need to register a feature map $\mathbf{F'}_{i}$ of a size $W'\times H'$, $\Delta_{ij}$ is bilinearly interpolated to that size, and shift values are scaled by the interpolation factors,
\begin{equation}
\label{eq:shiftMatrix2}
    \Delta' = bilinear(\Delta_{ij},\ H'\times W') \begin{bmatrix}
    W'/W & 0\\ 
    0 & H'/H
\end{bmatrix}.
\end{equation}
We equate the $k$-th element of  $\Delta'$ into Eq.~\ref{eq:valueReproj} to shift the feature elements to the desired frame as follows,
\begin{equation}
    \mathbf{F}^{*}_{j}(\mathbf{q}_{ik}+\Delta'_{k},c_k) \ = f_{jk}^{*} = f'_{ik} =\ \mathbf{F}'_{i}(\mathbf{q}_{ik},c_k) \\
\end{equation}
%

Due to perspective distortions when the camera moves, occlusions can make multiple re-projected elements fall in the same destination coordinate.
Since we want to predict masks in the current frame, we are only interested in points visible in it.
Therefore, we keep the point with the smallest registered depth since it  occludes any other re-projected ones in the current frame.

Fig.~\ref{fig:fusion} depicts this process for the $n$-th feature map of the decoder, where the block denoted with $\pi$ computes the shift matrix, which is down-sampled and employed to shift the desired feature map.
This process is implemented in a vectorized manner using PyTorch tensor operations to enable efficient execution on GPUs.
The implementation details of this are provided within the publicly available codebase (to be made available upon publication).
A downside of this approach is that the shifted feature maps will be noisy and potentially sparse. 
To overcome this issue we perform a fusion step as described below.

\begin{figure}[t!]
    \vspace{12pt}
    \centering
    \includegraphics[width=0.8\columnwidth]{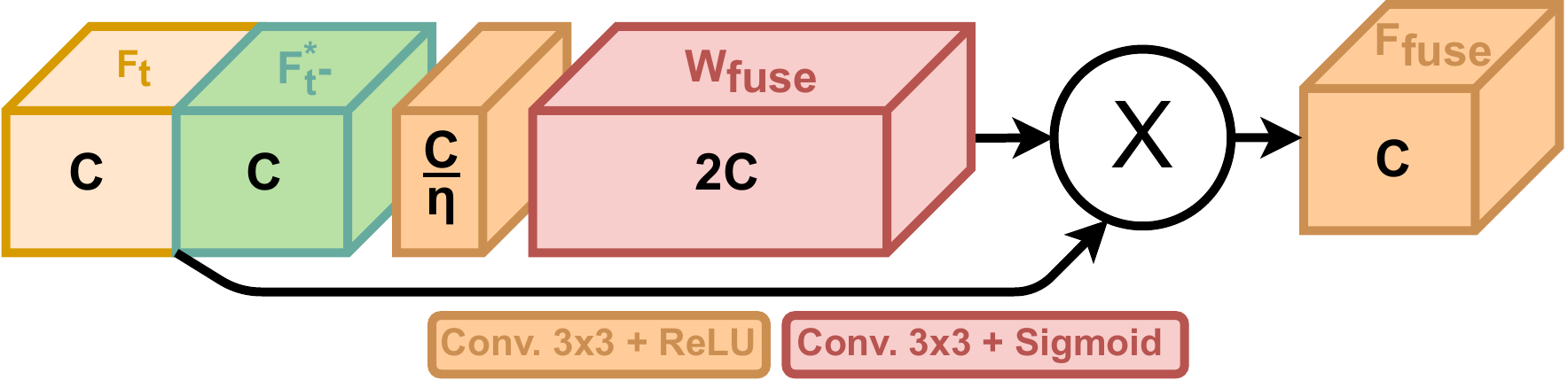}
    \caption{SSMA self-attention computational graph}
    \label{fig:SSMA}
    \vspace{-18pt}
\end{figure}

\subsection{Tensor fusion modules}
\label{sec:models}



We employ feature map/Tensor Fusion models (depicted in Fig.~\ref{fig:fusion}) for both RNNs and GRUs to explore the potential of short-term only as well as short- and long-term memory.
In both cases, we need to combine a resultant shifted feature map $\mathbf{F}^{*}_{t-}$ with the current feature map $\mathbf{F}_t$.
We cannot simply concatenate these two feature maps as the prior map is now potentially noisy and sparse.
This is because the depth measurement is imperfect and  was partially tackled with multi-modal attention in~\cite{valada2019self}.
This leads us to employ Tensor Fusion modules.
The aim of these modules is to extract and combine the most relevant information from the feature maps.



\subsubsection{Self-attention fusion}
\label{sec:selfAttention}

When fusing the two feature maps in an RNN we are inspired by  the prior work on multi-modal fusion by Valada et al.~\cite{valada2019self}.
We make use of the same attention mechansim, referred to as Self-Supervised Model Adaptation (SSMA), depicted in Fig.~\ref{fig:SSMA}.
This module computes a self-attention weight tensor $\mathbf{W}_{fuse}$ by concatenating the input feature maps $\mathbf{F}_{t}$ and the registered prior feature map $\mathbf{F}^{*}_{t^{-}}$, both consisting of $C$ channels.
The concatenated feature maps are then channel-wise compressed (bottleneck layer) with the aim being that only the important aspects will be maintained; the channel compression factor $\eta$ is a hyper-parameter of this model.
The resultant bottleneck is then channel-wise re-expanded to $2C$ using a sigmoid function which results in a map of values in the range $0$ to $1$.
This re-expanded feature map can then be considered to be the self-attention where $0$ indicates no attention should be given (e.g. from the prior or current feature map).
This learnable and data driven self-attention map ($\mathbf{W}_{fuse}$) is then Hadamard-multiplied with the concatenation of $\mathbf{F}_{t}$ and $\mathbf{F}^{*}_{t^{-}}$.
The final step is to channel-wise compress (bottleneck) this $2C$ matrix to size $C$ to obtain the final fusion map $\mathbf{F}_{fuse}$.
In this way, $\mathbf{F}_{fuse}$ can be fed back into the standard DCNN structure without further alteration.

A limitation of the above approach is that RNNs are only able to capture short-term temporal information.
Therefore, we also consider long-term short-term models (LSTMs)~\cite{hochreiter1997long} and how to perform spatial-temporal fusion with them.

\subsubsection{Conv-GRU fusion}
\label{sec:gru}

LSTMs allow the temporal models to exploit both long- and short-term temporal information.
This is achieved by maintaining hidden states which leads to a rapid expansion in the number of parameters in the model.
Recently it has been shown~\cite{bengio2014gru} that alternative formulations such as GRUs are able to perform at a similar level to LSTMs while incorporating fewer model parameters.

We use a GRU for temporal information and we deal with image feature maps by utilizing a ConvGRU~\cite{ballas2015delving}.
The ConvGRU takes as input 2D tensors (e.g. feature maps) and performs operations on these 2D tensors rather than the more standard vector representations.
This allows us to use them directly as a fusion mechanism in our proposed approach.
Namely, we provide as input to the GRU the current feature map $\mathbf{F}_{t}$ and this is then combined with the re-projected hidden state of the GRU.

\section{Experimental Setup}
\label{sec:experiments}

We conduct our experiments in two challenging agricultural scenarios: arable farming (crop vs weed) and horticulture (fruit segmentation).
Advantageously, these two scenarios provide a relatively static scene as the robot moves through. 
Yet, it is simultaneously challenging as the data is captured outdoors with no active illumination, and the scenes contain objects of vastly different sizes (e.g. growth stage) as well as high levels of occlusion.

The scene structure of these two scenarios are very different from relatively simple (arable farming) to highly complicated (horticulture).
In arable farming the object of interest is above the background (soil)  leading to a relatively simple scene structure. 
By contrast, the horticultural environment has a very cluttered scene with stem, branches, leaves and fruit providing occlusions from each other.
This enables us to draw deeper insights regarding where our proposed spatial-temporal models provide the greatest improvements.
More details about this data and our evaluation protocols are provided in Sec.~\ref{subsec:datasets}.

To incorporate the spatial-temporal information, we require an estimate of the robot's pose and depth information.
To obtain this information, we exploit the fact that both datasets were acquired on robots with a range of sensors as described in Sec.~\ref{subsec:wheel_odom}.
We also provide details about the baseline model and implementation details in Sec.~\ref{subsec:hyp_exp}.



%

\subsection{Datasets}
\label{subsec:datasets}
We train and evaluate the proposed models in 2 challenging agricultural scenarios, where accurate image segmentation is key to enable the automation of related processes.
The datasets we use are captured using RGB-D sensors and jointly captured a range of extra information such as motion information (e.g. odometry) which are critical for the deployment of our approaches. These datasets are, to the best of our knowledge, the only agriculture datasets with all modalities required by our approaches.
\begin{figure*}[!t]
    \vspace{12pt}
    \centering
    \includegraphics[width=\textwidth]{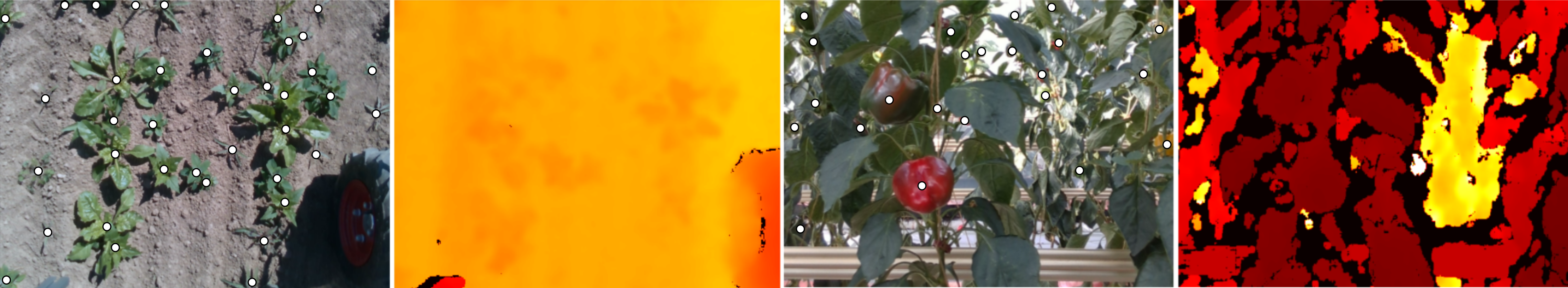}
    \caption{Datasets example images with instances marked with white dots. From left to right SB20 RGB (41 instances), SB20 depth, cropped BUP20 RGB (23 instances), and BUP20 depth.
    Brighter colors in the depth image represent farther objects.}
    \label{fig:dbexample}
    \vspace{-16pt}
\end{figure*}

\textbf{Arable Farming Sugar Beet (SB20)}, first introduced in~\cite{alireza2021Virtual}, was captured at a sugar beet field in campus Klein-Altendorf (CKA) of the University of Bonn using an Intel RealSense D435i camera with a nadir view of the ground mounted on BonnBot-I~\cite{ahmadi2021towards} driving at $0.4 m/s$. 
Sequences contain robot wheel odometry and RGB-D images of crops and 8 different categories of weeds at different growth stages, different illumination conditions and three herbicide treatment regimes ($30\%$, $70\%$, $100\%$), impacting weed density directly.
We evaluate models on a multi-class segmentation task for this dataset, only for its super-classes (crop, weed and background).

\textbf{Horticulture Glasshouse Sweet Pepper (BUP20)} consists of video sequences from a glasshouse environment in CKA, with two sweet pepper cultivar \textit{Mavera} (yellow) and \textit{Allrounder} (red), each cultivar matured from green, to mixed, to their primary color. 
This data, which contains all the colors, was captured by the autonomous phenotyping platform PATHoBot \cite{smitt2021} driving at $0.2 m/s$.
The dataset comprises 10 sequences of 6 crop-rows, captured using Intel RealSense D435i cameras recording RGB-D images, IMU data and wheel odometry.
These sequences were used to evaluate models in a binary-class segmentation task. 

Both datasets have sparse non-overlapping instance based segmentation annotations, allowing us to generate sequences of arbitrary length $N$, where only its last frame is labeled.
The resulting sequences provide challenging conditions with view-point perspective distortion, natural occlusions, motion blur, and illumination changes.
Unfortunately, we only have a single annotated frame for each sequence and this means that although we use all of the $N$ frames for training a loss is only produced for the last frame.
Therefore, models need to, at least, exploit temporal information to leverage all sequence frames for training.
A summary of attributes for these two datasets is provided in Tab.~\ref{tab:datasets}.
The relatively small number of annotated frames is compensated by the fact that there are a large number of instances in each frame (especially for SB20 which has high weed densities) as shown in the example images of Fig.~\ref{fig:dbexample} (both RGB and depth).

\begin{table}[!b]
    \vspace{-5mm}
	\centering
	\caption{Dataset Characteristics}
	\begin{tabular}{l ccc cc}
        & Image Size & fps & Train & Val. & Eval.\\\hline
    BUP20   & $704\times416$ & 15 & 124 & 62 & 93 \\
    SB20    & $640\times480$ & 15 & 71  & 37 & 35 \\
    \end{tabular}
	\label{tab:datasets}
\end{table}

\subsection{Frames Sequencing}
We train our models with sequences of $N$ frames as samples, and a batch is a set of these sequences.
These include RGB-D images and associated camera/robot odometry.
A limitation of temporal models is that they are generally trained and evaluated on consistent framerates. 
However, real-world systems need to deal with variable framerate as well as frame drops and jitter.
In this work we explicitly assess this issue and so use two frame sequence arrangements:

\textbf{Regular spacing} where $N$ consecutive frames from the dataset are fed as a sample to the model.

\textbf{Random spacing} consisting of $N$ frames with a random number of frameskips between them for each sample, with the $i$-th frame index $f_i$ computed by,
\begin{equation}
\label{eq:frameidx}
f_{i} = f_{i+1} - \delta, \ where \ 
\left\{\begin{matrix}
\delta \sim \mathcal{U}(1,\delta_{max}) \\ 
i \in [N-1, \cdots, 1]
\end{matrix}\right.
\end{equation}
where $\delta$ is a positive uniformly distributed variable with $\delta_{max}=6$, such that there is always a considerable overlap between frames in this extreme case.
This creates sequences with artificial frame drops and jitter with monotonously increasing indices, as well as longer camera baselines.
We train and evaluate all our models with $N=5$.

\subsection{Wheel dometry refinement}
\label{subsec:wheel_odom}

Our proposed spatial-temporal approach requires registration of feature maps.
This can be achieved in several ways and for this work, we choose to estimate the relative position change of the robot (e.g. camera), for registration, via both wheel and depth registration.
Wheel odometry is used as an initialization for RGB-D odometry, making use of Open-3D's ICP-based pipeline~\cite{open3d}.
We do this to account for wheel odometry errors produced by slippage, specially on SB20, and timestamp jitter with respect to camera frames. We explore the impact of this refinement in Sec.~\ref{sec:odometryAblation}.
Since our models infer form frames spatially close to each other, having locally consistent poses is enough, avoiding the need for globally consistent pose estimation methods.

\subsection{Implementation details}
\label{subsec:hyp_exp}

We perform semantic segmentation and chose U-Net as our base model to enable architecture fast-prototyping.
This is because of its simplicity in terms of layer operations, which requires little work to incorporate our spatial-temporal layers.
However, considering the variability in performance of segmentation architectures in agriculture~\cite{khan2020ced,hashemi2022deep} we also compare to DeepLabV3 on both datasets.

All of our networks are implemented in PyTorch along with the PyTorchLightning~\cite{falcon2019pytorch} wrapper to simplify infrastructure development.
We use Adam~\cite{kingma2014adam} as the training optimizer with a momentum of $0.8$, as well as a step-learning rate scheduler with a decay rate of $0.8$ decreasing its value every $100$ epochs and an initial learning rate value of $1e^{-3}$. 
Models were trained for at most $500$ epochs and validated every $5$ epochs.
Finally, the model with the best validation metric is chosen for evaluation.

Since our datasets are relatively small, we limited the training batch size to $6$.
However, some of our models (e.g. GRUs) require large amounts of memory to train, limiting the batch size to $4$.
All our models were trained from scratch on a server with A6000 GPUs with 48GB of GDDR6 memory and we made use of multi-GPU training to maintain batch sizes of $4$ and $6$ for our larger recurrent models.

\subsection{Weighted loss function and evaluation metrics}
Due to class imbalance present in the datasets, we employ a segmentation cross-entropy loss weighted according to inverse class frequencies as described in~\cite{milioto2020lidar}
\begin{equation}
\label{eq:loss}
    \mathcal{L} = - \sum\limits_{c=1}^{C} w_c\ y_c \log(\hat{y}_c),\ 
    \left\{\begin{matrix}
    w_c=1/\log(f_c+\epsilon) \\ 
    f_c = a_c / \sum_{i=1}^C a_i
    \end{matrix}\right.
\end{equation}
being $C$ the number of classes and $a_c$ the total area in pixels of class $c$ over the complete dataset.
This way, samples of classes with lower occurrence produce higher losses.

We use the intersection over union (IoU) percentage as our evaluation metric.
For SB20 we use the weighted IoU ($wIoU[\%]$) with the same class weights as presented in Eq.~\ref{eq:loss}.
As BUP20 is a binary segmentation task we use the mean IoU ($mIoU[\%]$).


\subsection{Evaluated models}

To understand the effectiveness of explicitly incorporating spatial-temporal cues, we compare against a range of models.
We use a still-image approach as a standard baseline and then analyze modified versions of this with only temporal information and with spatial-temporal information.
A summary of the models that we consider is given below.

\textbf{UNet Baseline (BL)}: A U-Net~\cite{ronneberger2015u} still-image semantic segmentation FCNN.

\textbf{DeepLabV3}: A DeepLabV3~\cite{chen2017rethinking} still-image semantic segmentation model with a ResNet50 backbone.

\textbf{Temporal GRU (T-GRU)}: Adds ConvGRU recurrent fusion modules (\ref{sec:gru}) at various depths to the segmentation decoder, thus incorporating temporal information.
Due to memory limitations, only 2 of these modules were employed, and early hyper-parameter search experiments found the best performance on BUP20 by fusing features $[F_{0}, F_{1}]$, and $[F_{1}, F_{2}]$ for SB20, which was adopted in all further experiments.
We implement this module with 2D convolutions with 3X3 learnable kernel filters, and their in/out sizes are determined by their input feature map.

\textbf{Spatial-Temporal GRU (ST-GRU)}: Uses feature map registering modules (\ref{sec:tensorReproj}), to pre-align its hidden state from previous frames to the ConvGRUs, thus explicitly incorporating both spatial and temporal information.

\textbf{Temporal self-attention (T-Atte)}: Employs SSMA as recurrent fusion modules (\ref{sec:selfAttention}) in its semantic decoder to incorporate temporal information.
Since this model has fewer parameters than our ConvGRUs, 3 of these modules were added, fusing features $[F_{0}, F_{1}, F_{2}]$ for BUP20, and $[F_{1}, F_{2}, F_{3}]$ for SB20, again determined by hyper-parameter search.
We set the compression factor of all these modules $\eta=16$, and all 2D convolution have 3x3 learnable kernels.

\textbf{Spatial-temporal self-attention (ST-Atte)}: This model also employs the feature registering module (\ref{sec:tensorReproj}) to exploit spatial information for the recurrent SSMA attention fusion module, making it also a spatial-temporal model.

\section{Results}
\label{sec:results}
\begin{table}[!b]
    \vspace{-4mm}
	\centering
	\caption{Results for training \& testing on random sequences.}
    \begin{tabular}{lcccc}

\toprule
\multicolumn{1}{c}{} & \multicolumn{2}{c}{BUP20} & \multicolumn{2}{c}{SB20} \\
Model & $mIoU[\%]$ & $\Delta_{BL}$ & $wIoU[\%]$ & $\Delta_{BL}$ \\
\midrule\hline
UNet (BL)            &   75.1            & -                             & 73.3          & -                             \\
DeepLabV3       &   73.6            & {\color{Red}$\downarrow 1.5$}                             & 74.6          & {\color{Green}$\uparrow 1.3$}                             \\\hline
T-GRU           &   75.0            & {\color{Red}$\downarrow 0.1$} & 76.2          & {\color{Green}$\uparrow 2.9$} \\
ST-GRU          &   77.5            & {\color{Green}$\uparrow 2.4$} & 75.3          & {\color{Green}$\uparrow 2.0$} \\
T-Atte          &   74.4            & {\color{Red}$\downarrow 0.7$} & 77.5          & {\color{Green}$\uparrow 4.2$} \\
ST-Atte         &   \textbf{77.7}   & {\color{Green}$\uparrow 2.6$} & \textbf{78.0}          & {\color{Green}$\uparrow 4.7$} \\
\bottomrule
\end{tabular}
\label{tab:results}
\end{table}

We perform three sets of quantitative experiments and finish with qualitative insights into the proposed approaches.
First, we summarize the evaluation results of all models trained and tested on random frame spacing as this is most reflective of a robot deployed in the real-world (e.g. simulating frame drop or inconsistent frame rates). 
Second, we analyze the robustness of the spatial-temporal and temporal models when trained and tested on different frame sequencing conditions.
Third, we assess our models' sensitivity w.r.t. estimated pose, by comparing results using wheel and RGB-D odometry.
Finally, we present a qualitative analysis on the two datasets and highlight the potential gains which are possible when deployed in challenging environments.

\subsection{Overall performance}

\begin{figure*}[ht!]
    \vspace{12pt}
    \centering
    \includegraphics[width=0.95\textwidth]{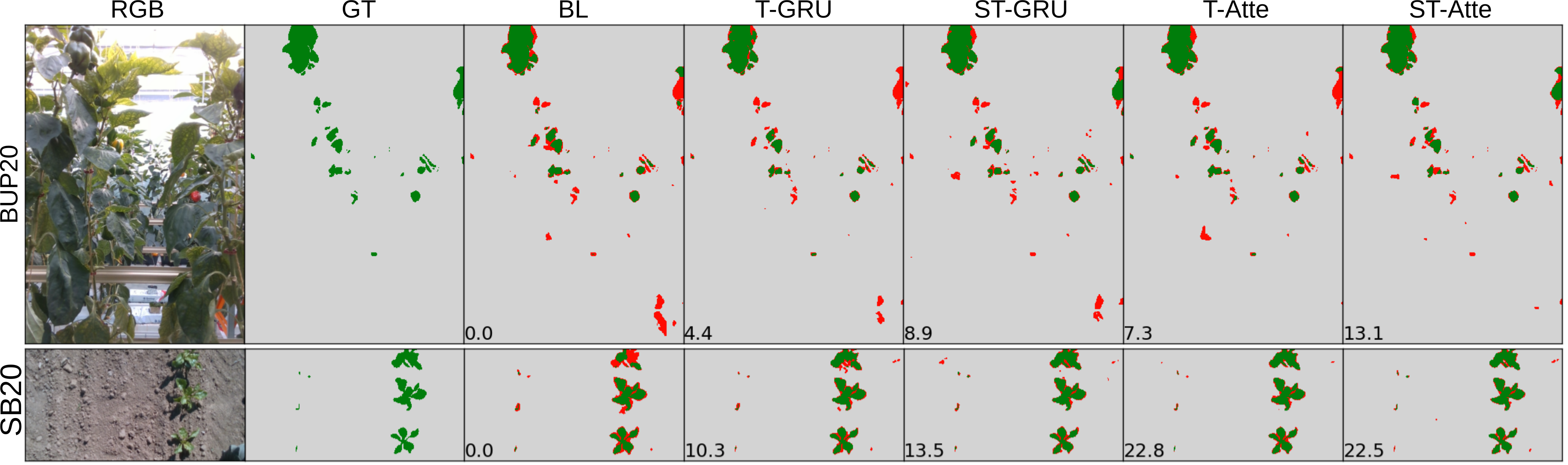}
    \caption{Qualitative results for BUP20 (top row) and SB20 (bottom row) for all of the models.
    We compare against the groundtruth (GT) where we highlight correct classifications in green and incorrect classification in red.
    For each model we present the IoU performance for the image compared to the baseline (BL).}
    \label{fig:serResults}
    \vspace{-17pt}
\end{figure*}


We summarize the evaluation results of all models trained and tested on random spacing frames in Tab.~\ref{tab:results}. 
We see that U-Net (our baseline) and DeepLabV3 have similar performance with U-Net achieving $1.5$ points higher on BUP20 and $1.3$ points lower on SB20.
Also, it can be seen that the spatial-temporal models (ST-Atte and ST-GRU) consistently outperform the baseline (U-Net) and DeepLabV3.
The best spatial-temporal model (ST-Atte) outperforms the baseline model by $4.7$ points on SB20 and $2.6$ points on BUP20. Furthermore, this model outperforms DeepLabV3 by $2.9$ points on SB20 and $3.9$ points on BUP20.
Thus, from now on we focus on comparing our proposed approaches with their baseline model to assess performance improvements.


Interestingly, both temporal-only models provide considerable gains, comparable to their spatial-temporal counterparts when there is a relatively uncluttered scene.
For example, this can be seen in SB20 where the temporal-only model (T-Att) has an absolute performance gain of $4.2$ points~and the spatial-temporal one (ST-Atte) presents a minor relative gain of $0.5$. 
Yet for more cluttered scenes such as BUP20 the performance actually degrades marginally by $0.7$ points; a similar trend also exists for T-GRU.
The lack of improvement for BUP20 is most likely due to the fact that there are frequent self-occlusions as well as large viewpoint changes. 


Finally, we observe the self-attention models tend to perform better than their GRU counterparts.
For instance, the ST-Atte model outperforms the ST-GRU model on both BUP20 and SB20 by $0.2$ points and $2.7$ points respectively.
This performance gain occurs despite the fact that the GRU model requires considerably more memory and computational power.
As such, we conclude that the ST-Atte model is the most appropriate for all assessed agricultural scenarios. 

\subsection{Robustness analysis to variable framerate}

\begin{table}[!b]
	\centering
	\caption{Variable framerate ablation study results for models trained on regular frames}
\begin{tabular}{lcccccc}
\toprule
{} & \multicolumn{3}{c}{BUP20 - $mIoU[\%]$} & \multicolumn{3}{c}{SB20 - $wIoU[\%]$} \\
Model  &        reg & rnd & $\Delta$ &         reg & rnd & $\Delta$ \\
\midrule\hline
T-GRU        & 76.3 &  72.9 &  {\color{Red}$\downarrow 3.4$} &   74.3 &    71.0 &    {\color{Red}$\downarrow 3.3$} \\
T-Atte       & 75.9 &  72.0 &  {\color{Red}$\downarrow 3.9$} &   76.7 &    75.0 &    {\color{Red}$\downarrow 1.7$} \\\hline
ST-GRU       & 76.6 &  76.2 &  {\color{Red}$\downarrow 0.4$} &   75.8 &    74.8 &    {\color{Red}$\downarrow 1.0$} \\
ST-Atte      & 75.3 &  74.6 &  {\color{Red}$\downarrow 0.7$} &   75.5 &    75.7 &    {\color{Green}$\uparrow 0.2$} \\
\bottomrule
\end{tabular}
\label{tab:ablation}
\end{table}

To better understand the performance of the spatial-temporal models we evaluate their performance in the presence of variable framerate, that is when the framerate for training is mismatched with the framerate at inference (testing) time.
Tab.~\ref{tab:ablation} summarizes the performance of the temporal and spatial-temporals in this scenario, where \textit{reg} indicates that the models are trained and tested with a regular framerate and \textit{rnd} indicates that the models are trained with a regular framerate and tested on a random framerate.


Our first observation is that the performance of the temporal-only models degrades considerably when the framerate is mismatched.
The most robust temporal model, T-Atte, degrades by $3.9$ points and $1.7$ points on BUP20 and SB20 respectively.
By contrast, the spatial-temporal modes are considerably more robust.

When spatial-temporal models are trained and tested in miss-matched conditions they have only minor performance degradation. 
For the best performing ST model, ST-Atte, there is a minor performance degradation of $0.7$ points for BUP20 and a minor performance improvement of $0.2$ points on SB20.
Similarly, the ST-GRU approach only degrades by $0.4$ points on BUP20 and $1.0$ points on SB20.
This shows that, despite being trained on regular frames, the ST models are more robust to variable framerate sequences.
We attribute this to their ability to garner extra information from the prior spatial information. 
This leads us to conclude that spatial information is crucial for segmentation models robust to variable framerates, which is a relevant issue when operating in complex agricultural environments.

\subsection{Robustness to odometry errors}
\label{sec:odometryAblation}

\begin{table}[!b]
    \vspace{-4mm}
	\centering
	\caption{Robustness to odometry errors results}
\setlength\tabcolsep{4pt} 
\begin{tabular}{lcccccc}
\toprule
                         &       \multicolumn{3}{c}{BUP20 - $mIoU[\%]$}         & \multicolumn{3}{c}{SB20 - $wIoU[\%]$} \\
Model                    &       Wheel & RGB-D & $\Delta$                       & Wheel & RGB-D & $\Delta$  \\
\midrule\hline
ST-GRU & 77.2  & 77.5  & {\color{Green}$\uparrow 0.3$}   & 74.1  & 75.3  & {\color{Green}$\uparrow 1.2$}   \\
ST-Atte  & 77.3  & 77.7  & {\color{Green}$\uparrow 0.4$}   & 75.3  & 78.0  & {\color{Green}$\uparrow 2.7$}   \\                   
\bottomrule
\end{tabular}
\label{tab:odometryAblation}
\end{table}

An important element in our spatial-temporal models is the estimated camera pose to perform the spatial registration of feature maps (see Sec.~\ref{sec:tensorReproj}).
So far we showed results using RGB-D refined odometry which provides accurate camera poses.
To evaluate the robustness of our models to pose errors we compare these results to employing raw wheel odometry information instead.
The results in Tab.~\ref{tab:odometryAblation} highlight the importance of accurate odometry information.
For BUP20 using RGB-D odometry, rather than wheel odometry, results in modest performance gains of $0.3$ and $0.4$ points for ST-GRU and ST-Atte respectively, whereas on SB20 the impact is more evident with improvements of $1.2$ and $2.7$ respectively.
We attribute the difference in performance gains, to the environmental differences. 
In BUP20, the robot is in a constrained environment on pipe rails and so wheel slippage leads to minimal odometry errors.
By comparison, in SB20, the robot is outdoors in a field and consequently much larger slippage errors are possible.

Despite the impact of odometry refinement on the proposed approach, even wheel odometry is sufficient to provide an improved system.
For BUP20, wheel odometry alone leads to an improvement over the baseline of $2.1$ and $2.2$ points for ST-GRU and ST-Atte respectively.
While for SB20, wheel-odometry alone leads to an improvement over the baseline of $0.8$ and $2.0$ for ST-GRU and ST-Atte respectively.

\subsection{Segmentation output results}

In Fig.~\ref{fig:serResults} we present qualitative results on all of the approaches.
It can be seen that ST-Atte provides high-quality segmentation outputs with mostly minor errors.
For SB20, the most controlled scenario, it can be seen that the bulk of the errors are minor segmentation errors in larger plants, however, it provides considerably better results than the baseline (BL). 
For BUP20, the most challenging scenario, it can be seen that the baseline and GRU models (T-GRU and ST-GRU) consistently mistake a foreign object (red tape) for sweet pepper. 
However, both RNN models (T-Atte and ST-Atte) are able to correctly classify this region.
The performance improvements for ST-Atte, over T-Atte, can be seen by more accurate segmentation and fewer spurious segmentation results (both positive and negative).
Furthermore, both spatial-temporal models can correctly identify the existence of sweet pepper in the border of the images (top right) which every other system incorrectly classifies.


\section{Summary}

In this paper, we have explored approaches to explicitly capture both spatial and temporal information to improve the classification of DCNNs. 
We have shown that RNNs and GRUs can leverage available RGB-D images and robot odometry information to improve segmentation performance. 

Our experiments were performed on two challenging agricultural datasets in arable farming (crop vs weed) and horticulture (fruit segmentation).
On these datasets, we demonstrated that our approaches can considerably improve the classification performance and our best performing spatial-temporal model (ST-Atte) achieves an absolute performance improvement of $4.7$ for crop-weed segmentation and $2.6$ for fruit (sweet pepper) segmentation.

Additionally, we show that these spatial-temporal approaches are robust to variable framerates and odometry errors, as they suffer from only minor performance degradation in these conditions.
%
%
We conclude that the proposed spatial-temporal models provide consistently improved performance.
Finally, while we currently only apply these approaches in agriculture, we believe this technique has the potential to be exploited for 2D and 3D object segmentation in other domains, and optimized to run in real-time.

\bibliography{references}

\bibliographystyle{IEEEtran}


\end{document}